\documentclass[letterpaper,journal]{IEEEtran}

\usepackage{amsmath,amsfonts,amssymb}
\usepackage{array}
\usepackage[caption=false,font=footnotesize]{subfig}
\usepackage{textcomp}
\usepackage{stfloats}
\usepackage{placeins}
\usepackage{url}
\usepackage{graphicx}
\usepackage{cite}
\usepackage{booktabs}
\usepackage{tabularx}
\usepackage{makecell}
\usepackage{xcolor}
\usepackage{tikz}
\usepackage[ruled,vlined]{algorithm2e}
\usepackage{kotex}
\usetikzlibrary{arrows.meta,positioning,fit,calc}
\renewcommand{\abstractname}{Abstract}
\renewcommand{\IEEEkeywordsname}{Index Terms}

\newcommand{\figplaceholder}[1]{%
\fbox{\begin{minipage}[c][0.20\textheight][c]{0.95\linewidth}\centering #1\end{minipage}}}
\begin{document}

\title{Bridging the Sim-to-Real Gap in Parallel-Link Leg Mechanisms via Simulator-Side Dynamics Normalization}

\author{
Jinsong Hong,
Jangho Kim,
Jihwan Lee,
Donghyun Kim,
and Sehoon Oh

\thanks{This work was supported by the National Research Foundation of Korea
(NRF) grant funded by the Korea government (MSIT)
(No. RS-2024-00354028), by the Basic Science Research Program through the
National Research Foundation of Korea (NRF) funded by the Ministry of
Education (No. RS-2025-25420118), and by the InnoCORE Program of the
Ministry of Science and ICT (No. 26-InnoCORE-01).}%

\thanks{Jinsong Hong, Jihwan Lee, and Sehoon Oh are with the
Department of Robotics and Mechatronics Engineering, DGIST,
Daegu 42988, Republic of Korea
(e-mail: hn05198@dgist.ac.kr; fist5678@dgist.ac.kr;
sehoon@dgist.ac.kr).
Corresponding author: Sehoon Oh
(phone: +82-53-785-6209).}

\thanks{Jangho Kim is with the Artificial Intelligence Major,
DGIST, Daegu, Republic of Korea
(e-mail: jangho123.dgist@dgist.ac.kr).}

\thanks{Donghyun Kim is with the Manning College of Information and Computer Sciences, University of Massachusetts Amherst, Amherst, MA, USA (e-mail: donghyunkim@cs.umass.edu).}

\thanks{\textit{This work has been submitted to the IEEE for possible publication. Copyright may be transferred without notice, after which this version may no longer be accessible.}}

}

\maketitle

\begin{abstract}
This paper addresses the sim-to-real gap in dynamics arising when a parallel-link mechanism is represented by a serial-tree surrogate in simulation. Conventional Jacobian-based state and torque mappings preserve consistency with the kinematic and virtual-work relations but do not account for the coordinate-induced redistribution of actuator inertia and damping and the linkage inertia omitted during serial-tree reduction. To address this gap, Simulator-Side System Normalization (S3N) is proposed to normalize the serial-tree simulator's effective dynamics while preserving its tree topology. S3N-Act incorporates actuator inertia and damping into the serial-coordinate dynamics through coordinate transformation, whereas S3N-Full restores residual linkage inertia by separately identifying actuator- and leg-level frequency responses. In the 2-DoF validation, S3N-Full reduced the joint-position and torque RMSEs by 80.9\% and 82.1\%, respectively, relative to the Jacobian-mapping baseline. During pitch-in-place motion, S3N-Act and S3N-Full reduced the RMSE of the ground reaction force norm by 65.1\% and 62.4\%, respectively. During circular locomotion, S3N-Full reduced the phase-averaged, command-normalized sim-to-real gap from 17.3\% to 9.9\%. These results show that simulator-side normalization improves motion- and force-level sim-to-real consistency. It enables policy training in a serial-tree framework with hardware-consistent dynamics that better represent the physical parallel-link mechanism.

\end{abstract}


\section{Introduction}
\IEEEPARstart{P}{arallel} mechanisms provide high structural stiffness and load-carrying capacity through multiple load paths \cite{Merlet2006ParallelRobots,Briot2015ParallelRobotDynamics}. Applied to robotic legs, these mechanisms support the generation of large ground reaction forces through their load-sharing structure and mechanical advantage \cite{Oh2011BiarticularForceTransmission,Guo2017Pantograph,Wensing2017Cheetah}. Furthermore, the proximal placement of actuators and linkage-based transmission to the distal joints reduce distal leg mass and inertia, which is advantageous for rapid swing motions \cite{Wensing2017Cheetah,Hoffman2024KangarooHybridParallel}. These characteristics are well suited to dynamic quadrupedal locomotion, which requires rapid stance--swing transitions, repeated ground impacts, and large contact forces, and have motivated the adoption of parallel-link mechanisms in various legged robots \cite{Hong2024slip,Wensing2017Cheetah,Guo2017Pantograph}.

Simulation-based reinforcement learning has become a dominant approach for controlling legged robots. To exploit these mechanical advantages in learning-based control, simulation must reproduce not only the kinematic constraints of the closed-loop transmission structure but also its key dynamic responses \cite{Briot2015ParallelRobotDynamics,Tanaka2025ParallelHumanoid,Zhang2025HighDynamic,Guan2024ImpedanceMatching}. Otherwise, the policy experiences simulator dynamics that do not match the hardware dynamics, potentially leading to sim-to-real gaps in motion and contact-force responses. Approaches to representing parallel mechanisms in simulation can be broadly classified into constraint-based closed-chain modeling (CBCM) and transformation-based closed-chain modeling (TBCM). CBCM preserves the actual closed-loop topology by implementing loop-closure as constraints within the simulator. Its applicability has also been demonstrated in policy learning and hardware transfer \cite{Todorov2012MuJoCo,Amadio2025Kangaroo,Tanaka2025ParallelHumanoid}. However, in such a constraint formulation, loop-closure accuracy and dynamic response depend on the constraint and solver settings. Because these numerical settings do not directly correspond to the robot’s physical parameters, appropriate values are difficult to determine solely from hardware characteristics \cite{MuJoCoComputation,MuJoCoModeling}. By contrast, TBCM reduces the closed-loop mechanism to a serial-tree surrogate and transforms states and torques between the hardware-side parallel actuation coordinates and the simulation-side serial joint coordinates using coordinate relations and Jacobian-based virtual-work relations \cite{Ficht2017NimbRoOP2,Zhang2025HopParallel}. This approach retains existing tree-based rigid-body simulation and learning pipelines without explicit loop-closure constraints~\cite{Rudin2022CatLike,Zhang2025HopParallel,Wang2025BoosterGym}. Accordingly, this study focuses on improving the dynamic fidelity of the serial-tree simulator while preserving the TBCM framework.

Early examples of TBCM introduced a virtual serial leg to reuse existing serial-chain motion-control frameworks on parallel-link hardware. At the hardware interface, virtual serial-joint commands were converted into parallel-actuator commands, and actuator feedback was mapped back to the virtual serial coordinates \cite{Ficht2017NimbRoOP2}. Subsequent reinforcement-learning studies either trained policies using serial-tree surrogates and applied serial-parallel state and torque transformations during deployment \cite{Rudin2022CatLike,Wang2025BoosterGym}, or incorporated these transformations into the simulation control loop \cite{Zhang2025HopParallel}. Another approach approximates the omitted closed-chain linkage with a high-stiffness virtual spring \cite{Radosavovic2024Humanoid}. A different approach reconstructs the kinematic states of the parallel actuators and linkage from serial ankle motion at each simulation step and transforms parallel-space control torques for application in the serial simulator \cite{Zhang2025LiPS}. To capture nonlinear transmission and reduce the computational burden of iterative numerical solutions for closed-chain kinematics, another approach formulates the configuration-dependent joint-actuator transmission as an analytical actuation model \cite{Lutz2026DifferentialActuation}.

However, in conventional TBCM, mapping the states and torques of a parallel mechanism to a serial-tree surrogate does not recover the redistribution and coupling of actuator inertia and damping induced by the parallel-to-serial coordinate transformation or the linkage inertia omitted during serial-tree reduction. Approaches to restoring these parallel-mechanism dynamics in the serial-tree simulator so that the policy can experience the key inertial and damping responses of the physical mechanism during training have not been sufficiently explored. The model mismatch resulting from the omission of these dynamics can lead to sim-to-real gaps in joint acceleration, contact-force, and body-motion responses \cite{Kumar2019ModelSimplification}. Domain randomization can improve policy robustness to model uncertainty~\cite{Peng2018DynamicsRandomization,Kumar2021RMA,
Tao2026EfficientSim2Real}, but it does not explicitly restore the missing coupling and inertia structure in the simulator dynamics.

To address this limitation, this paper proposes Simulator-Side System Normalization (S3N), which restores the key inertial and damping responses of the physical parallel-link mechanism in a serial-tree simulator. S3N-Act transforms parallel-actuator inertia and damping into equivalent terms in serial joint coordinates and restores their redistribution omitted during the coordinate transformation. S3N-Full further identifies actuator- and leg-level frequency responses separately. S3N-Full then subtracts the actuator contribution from the identified total leg inertia to prevent double counting and estimates and restores the residual linkage inertia omitted during serial-tree reduction. S3N is implemented in an Isaac-based serial-tree learning environment \cite{NVIDIAIsaacSim510}. 

\begin{figure}[t]
    \centering
    \includegraphics[width=0.9\linewidth]{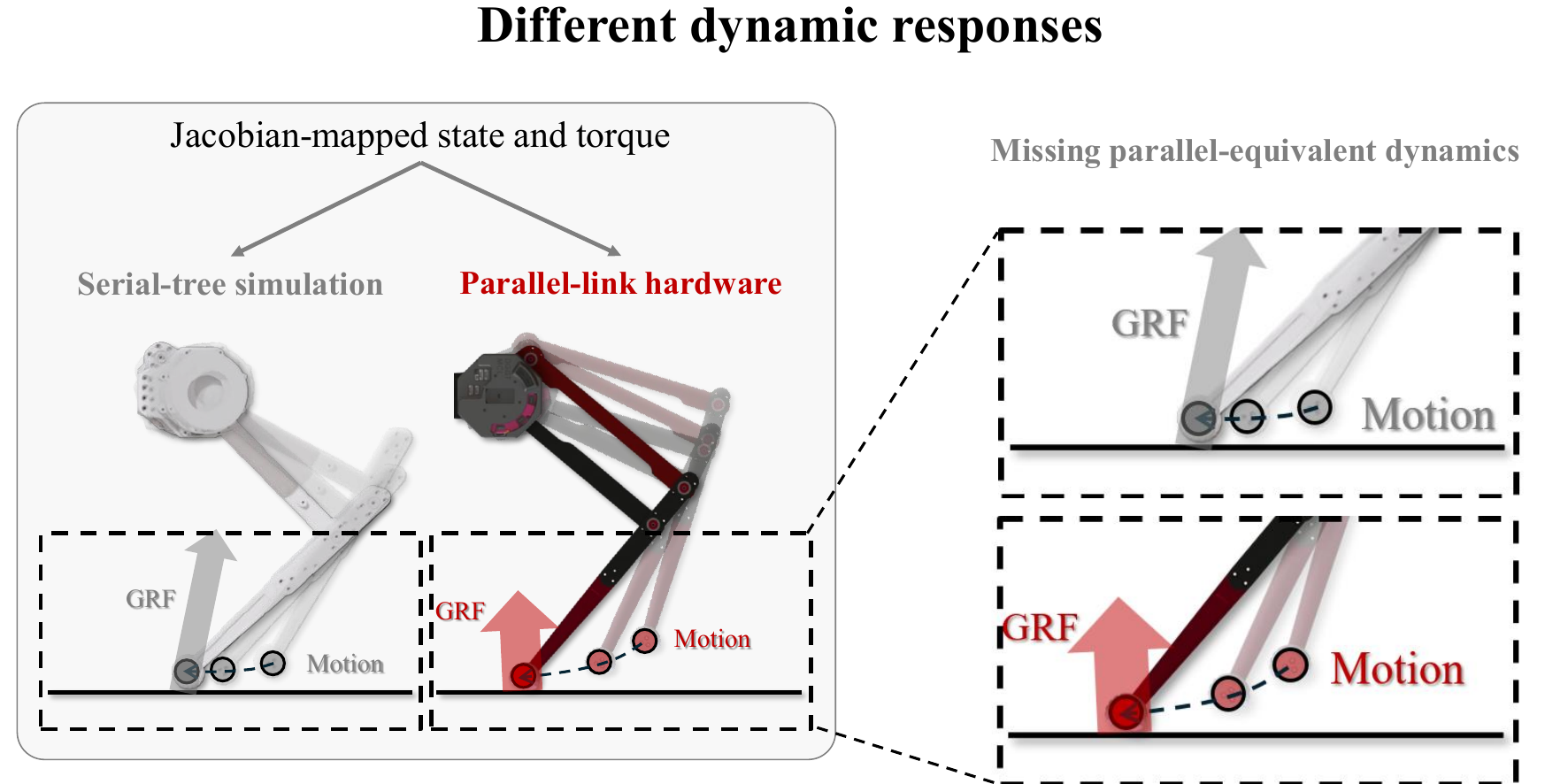}
    \caption{Problem formulation for a parallel-link mechanism modeled by a serial-tree surrogate. Jacobian-based state and torque mappings preserve the virtual-work relationships but not the closed-chain dynamics, leading to sim-to-real discrepancies in motion and GRF.}
    \label{fig:problem}
\end{figure}

The main contributions of this paper are as follows:
\begin{itemize}
\item The sim-to-real gap in dynamics caused by serial-tree reduction is identified and formalized by decomposing it into two components: the coordinate-induced redistribution and coupling of parallel-actuator inertia and damping, and residual linkage inertia.

\item A hardware-informed construction procedure is presented to identify the S3N normalization terms from measured actuator- and leg-level frequency responses and prevent double counting of actuator inertia.

\item The effectiveness of S3N in reducing motion- and force-level sim-to-real gaps is validated on a parallel-link quadruped using contact-free 2-DoF leg motion and torque responses, ground-reaction-force responses under fixed stance, and locomotion-policy transfer.
\end{itemize}

\section{Problem Statement}
\label{sec::problem}

We refer to the conventional transformation-based approach as the Kin-Only approach, because it uses only the kinematic relations—coordinate transformation and Jacobian-based virtual-work mapping—to convert between the parallel actuation coordinates and the serial surrogate, without compensating for the dynamic discrepancy introduced by the structural reduction. This section formulates the Kin-Only approach and identifies the resulting structure-induced dynamics gap.

\begin{figure}[!t]
    \centering
    \includegraphics[width=0.9\columnwidth]{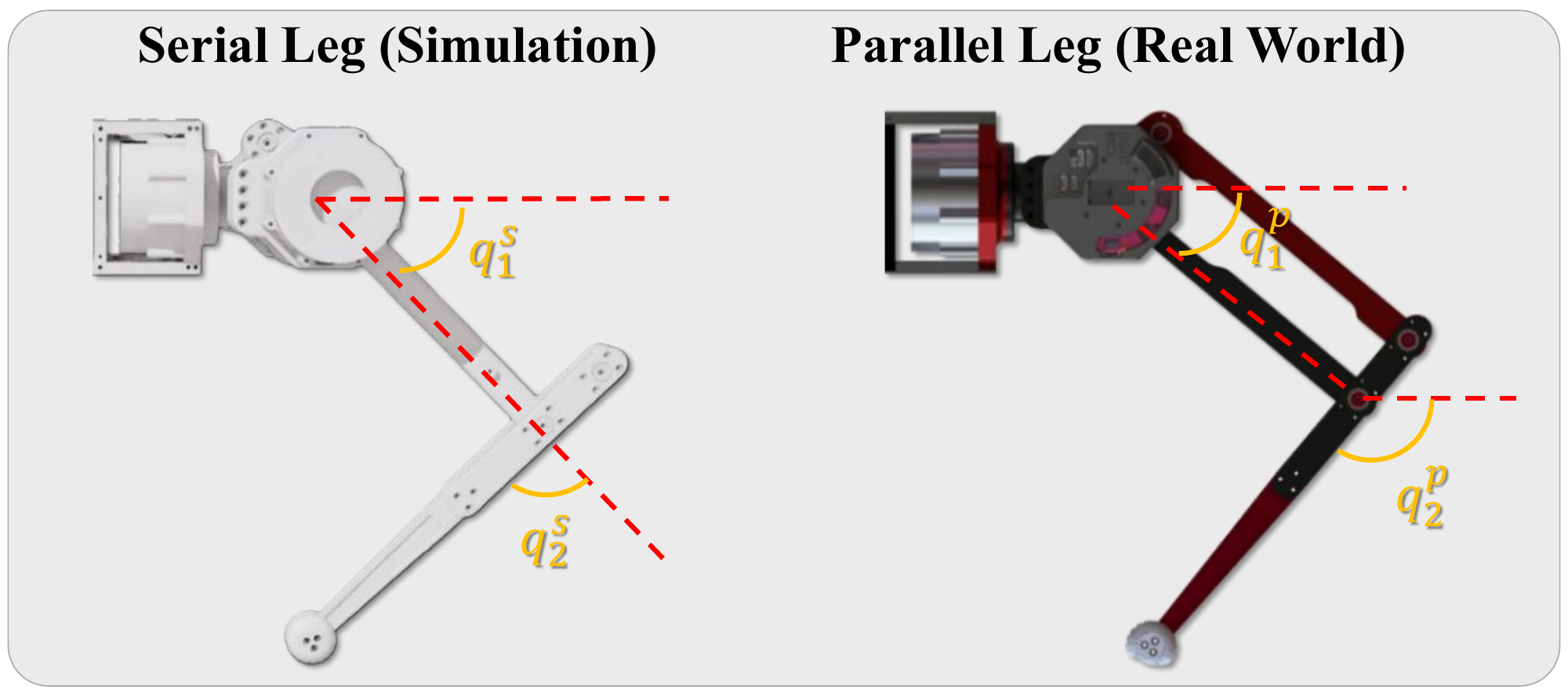}
    \caption{Coordinate definitions for the serial-tree simulator and the physical parallel-link leg.}
    \label{fig:leg_coordinate}
\end{figure}

Let the serial coordinates used by the simulator be \(\mathbf{q}^s=[q_1^s,q_2^s]^T\), and the physical parallel actuation coordinates be \(\mathbf{q}^p=[q_1^p,q_2^p]^T\). The relationship between the two coordinate systems is given by 
\begin{equation}
\mathbf{q}^p=\mathbf{J}\mathbf{q}^s,\quad \dot{\mathbf{q}}^p=\mathbf{J}\dot{\mathbf{q}}^s,\quad
\ddot{\mathbf{q}}^p=\mathbf{J}\ddot{\mathbf{q}}^s,
\label{eq:coordinate_relation}
\end{equation}
\begin{equation}
\mathbf{J}=
\begin{bmatrix}
1&0\\
1&1
\end{bmatrix}.
\label{eq:coordinate_jacobian}
\end{equation}
Accordingly, \(q_1^p=q_1^s\) and \(q_2^p=q_1^s+q_2^s\), and the relative angle between the two links in the physical parallel coordinates is expressed as
\begin{equation}
\alpha=q_2^p-q_1^p=q_2^s
\label{eq:relative_angle_alpha}
\end{equation}
The generalized torques satisfy the following virtual-work relation \cite{Lynch2017ModernRobotics}.
\begin{equation}
\boldsymbol{\tau}^s=\mathbf{J}^T\boldsymbol{\tau}^p.
\label{eq:torque_pullback}
\end{equation}
The coordinate relationship illustrated in Fig.~\ref{fig:leg_coordinate} determines not only the transformations of position and torque commands, but also how inertia, damping, and generalized forces should be redistributed in the serial coordinates.

In the Kin-Only approach, the mappings in Eqs.~\eqref{eq:coordinate_relation}--\eqref{eq:torque_pullback} are used to apply commands expressed in the parallel coordinates to the serial simulator. Accordingly, the serial simulator integrates the following joint-space dynamics:
\begin{equation}
\mathbf{M}_{s}(\mathbf{q}^s)\ddot{\mathbf{q}}^s+
\mathbf{h}_{s}(\mathbf{q}^s,\dot{\mathbf{q}}^s)+
\mathbf{D}_{s}\dot{\mathbf{q}}^s+
\mathbf{g}_{s}(\mathbf{q}^s)
=
\mathbf{J}^T\boldsymbol{\tau}^{p}.
\label{eq:serial_plant}
\end{equation}
Here, \(\mathbf{M}_{s}\), \(\mathbf{h}_{s}\), \(\mathbf{D}_{s}\), and \(\mathbf{g}_{s}\) denote the inertia, velocity-dependent rigid-body, damping, and gravity terms, respectively, of the serial-tree surrogate.

However, Jacobian mapping transforms only the coordinate and torque-command relationships. It does not recover the actuator dynamics redistributed by the coordinate transformation of the parallel actuation system and the linkage dynamics omitted from the serial-tree surrogate. In this study, the dominant components of the resulting structure-induced dynamics gap are expressed in the serial coordinates as follows.
\begin{equation}
\label{eq:problem_gap_decomposition}
\begin{gathered}
\Delta\mathbf{M}^{s}(\alpha)
=
\Delta\mathbf{M}_{\mathrm{act}}
+
\Delta\mathbf{M}_{\mathrm{link}}(\alpha),
\\
\Delta\mathbf{D}^{s}
=
\Delta\mathbf{D}_{\mathrm{act}}.
\end{gathered}
\end{equation}
Here, \(\Delta\mathbf{M}_{\mathrm{act}}\) and \(\Delta\mathbf{D}_{\mathrm{act}}\) denote the missing redistribution components that arise when the inertia and damping of the parallel actuators are transformed into the serial coordinates, while \(\Delta\mathbf{M}_{\mathrm{link}}(\alpha)\) accounts for both the residual linkage inertia omitted from the serial-tree surrogate and the discrepancy between the CAD-based and identified link inertias. Consequently, even when the same mechanism state and the corresponding generalized torques are applied, the Kin-Only model may cause the simulator and hardware to exhibit different joint accelerations, motions, and contact-force responses because of the missing inertia and damping components (Fig.~\ref{fig:problem}). In Sec.~\ref{sec:s3n}, each dynamics gap is derived analytically, and S3N is presented to incorporate these components into the serial simulator.

\section{Simulator-Side Dynamics Normalization for Hardware-Consistent Training}
\label{sec:s3n}

This section presents S3N to compensate for the structure-induced dynamics omitted in the Kin-Only formulation while retaining the Jacobian-based coordinate and torque mappings defined in the preceding section. First, we derive how the inertia and damping of the parallel actuators are redistributed in the serial coordinates and define S3N-Act to restore these components. Next, we construct the residual inertia of the parallel linkage assembly relative to the serial-tree surrogate from hardware identification results and present S3N-Full, which incorporates this linkage contribution in addition to the actuator-side normalization.

\begin{figure*}[t]
    \centering
    \IfFileExists{Figure/WholeFramework_fin.pdf}{\includegraphics[width=0.9\textwidth,height=0.39\textheight,keepaspectratio]{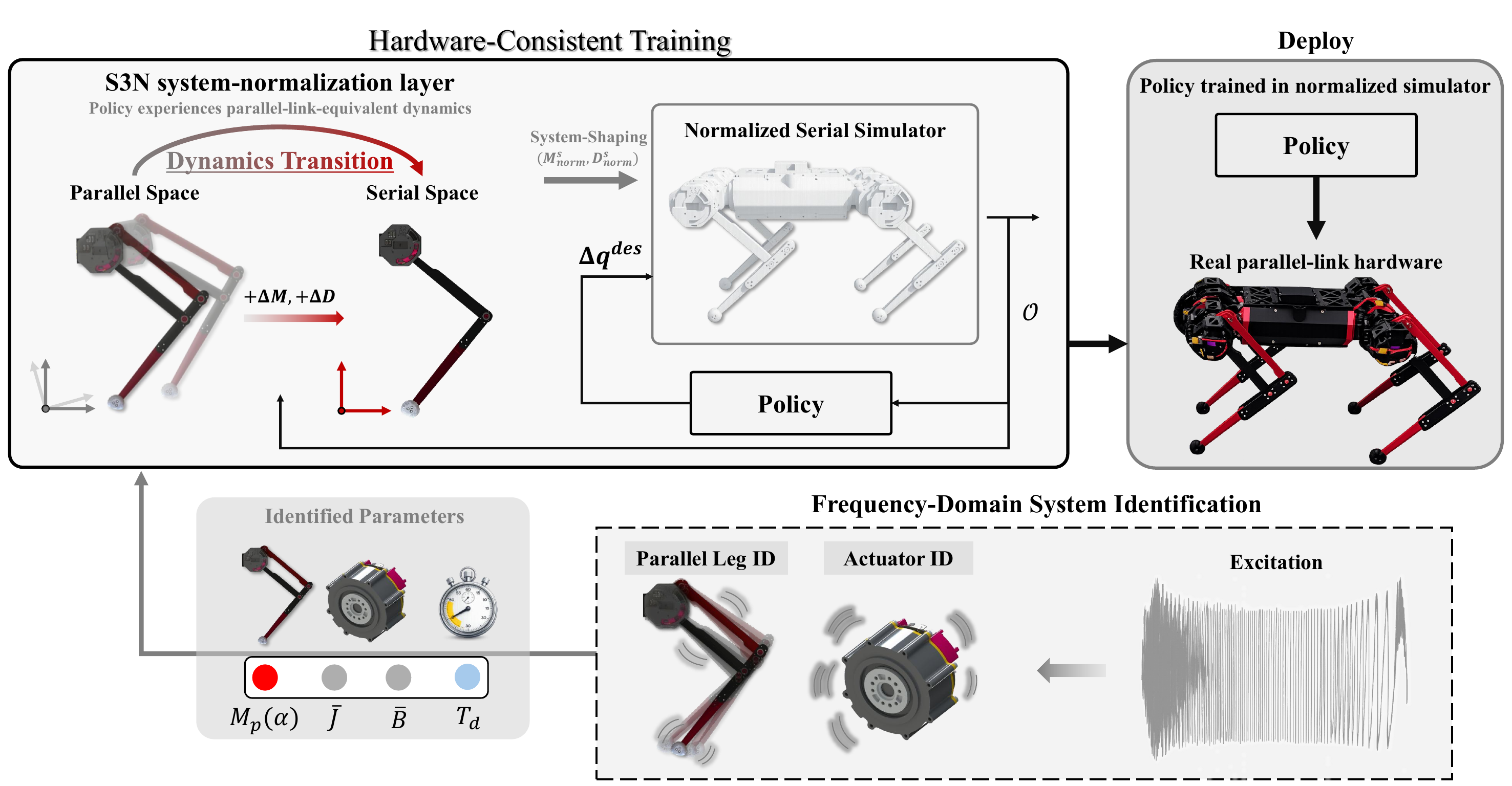}}{\figplaceholder{Figure/WholeFramework\_fin.pdf}}
    \caption{Simulator-side dynamics normalization with S3N. All variants retain the serial-tree simulator and use the same coordinate mapping. S3N-Act additionally restores the pulled-back actuator inertia and damping, whereas S3N-Full further adds residual linkage-inertia shaping. The shaping signal is used only inside the simulator and is neither a policy action nor a hardware command.}
    \label{fig:s3n_overview}
\end{figure*}

\subsection{Actuator-Side Inertia and Damping Normalization}
The motor inertia and viscous damping reflected to the actuator output side through the gear ratio \(N_i\) are denoted by \(\bar{J}_i\) and \(\bar{B}_i\), respectively.
\begin{equation}
\bar{J}_i=N_i^2J_{mi},\qquad
\bar{B}_i=N_i^2B_{mi}.
\label{eq:reflected_actuator_dynamics}
\end{equation}

Using the coordinate mapping in Eqs.~\eqref{eq:coordinate_relation}--\eqref{eq:relative_angle_alpha}, the actuator inertia and damping in the parallel coordinates are expressed as
\begin{equation}
\mathbf{M}_{\mathrm{act}}^p=
\begin{bmatrix}
\bar{J}_1&0\\
0&\bar{J}_2
\end{bmatrix},
\qquad
\mathbf{D}_{\mathrm{act}}^p=
\begin{bmatrix}
\bar{B}_1&0\\
0&\bar{B}_2
\end{bmatrix}
\label{eq:parallel_actuator_diag}
\end{equation}
Applying the inertia-coordinate transformation implied by kinetic-energy invariance \cite{Khatib1987OperationalSpace}, these matrices are pulled back to the serial-simulator coordinates as
\begin{equation}
\mathbf{M}_{\mathrm{act}}^{p\rightarrow s}
=
\mathbf{J}^T\mathbf{M}_{\mathrm{act}}^p\mathbf{J}
=
\begin{bmatrix}
\bar{J}_1+\bar{J}_2&\bar{J}_2\\
\bar{J}_2&\bar{J}_2
\end{bmatrix},
\label{eq:M_act_pullback}
\end{equation}
\begin{equation}
\mathbf{D}_{\mathrm{act}}^{p\rightarrow s}
=
\mathbf{J}^T\mathbf{D}_{\mathrm{act}}^p\mathbf{J}
=
\begin{bmatrix}
\bar{B}_1+\bar{B}_2&\bar{B}_2\\
\bar{B}_2&\bar{B}_2
\end{bmatrix}
\label{eq:D_act_pullback}
\end{equation}
In contrast, the actuator-side inertia and damping in the serial simulator are independently assigned to each serial joint and are therefore represented by the following diagonal matrices:
\begin{equation}
\mathbf{M}_{\mathrm{act}}^s=
\begin{bmatrix}
\bar{J}_1&0\\
0&\bar{J}_2
\end{bmatrix},
\qquad
\mathbf{D}_{\mathrm{act}}^s=
\begin{bmatrix}
\bar{B}_1&0\\
0&\bar{B}_2
\end{bmatrix}.
\label{eq:serial_actuator_diag}
\end{equation}
Therefore, the joint-space inertia and damping gaps required to obtain the target actuator-side normalized dynamics are
\begin{equation}
\Delta\mathbf{M}_{\mathrm{act}}
=
\mathbf{M}_{\mathrm{act}}^{p\rightarrow s}
-
\mathbf{M}_{\mathrm{act}}^s
=
\begin{bmatrix}
\bar{J}_2&\bar{J}_2\\
\bar{J}_2&0
\end{bmatrix},
\label{eq:Mact_delta}
\end{equation}
\begin{equation}
\Delta\mathbf{D}_{\mathrm{act}}
=
\mathbf{D}_{\mathrm{act}}^{p\rightarrow s}
-
\mathbf{D}_{\mathrm{act}}^s
=
\begin{bmatrix}
\bar{B}_2&\bar{B}_2\\
\bar{B}_2&0
\end{bmatrix}
\label{eq:Dact_delta}
\end{equation}
S3N-Act adds the inertia and damping increments derived in Eqs.~\eqref{eq:Mact_delta}--\eqref{eq:Dact_delta} to the simulated dynamics, thereby compensating for the components omitted by the Jacobian-based transformation. This adjusts the actuator-side dynamic response toward that of the physical parallel mechanism while preserving the existing serial-tree topology.

\subsection{Residual Parallel-Link Inertia Normalization}

Even after S3N-Act compensates for the actuator inertia and damping omitted during the coordinate transformation, an additional inertia mismatch remains between the physical parallel-link assembly and the serial-tree surrogate. This mismatch arises from both the linkage inertia omitted by the serial-tree approximation and the discrepancy between the CAD-based and hardware-identified values. In this section, we define and compensate for this mismatch as the residual linkage inertia.

First, the total inertia of the physical parallel-link leg is parameterized as
\begin{equation}
\mathbf{M}_{p}(\alpha)
=
\begin{bmatrix}
M_{11}&M_{12}\cos\alpha\\
M_{12}\cos\alpha&M_{22}
\end{bmatrix}.
\label{eq:Mp_structure}
\end{equation}
The \(\cos\alpha\) term arises from the kinetic-energy coupling of the planar four-bar linkage mechanism \cite{Briot2015ParallelRobotDynamics,Lynch2017ModernRobotics}. 

Subtracting the actuator-side inertia from the total inertia gives the link-side inertia in the parallel coordinates as
\begin{equation}
\mathbf{M}_{\mathrm{link}}^{p}(\alpha)
=
\mathbf{M}_{p}(\alpha)
-
\mathbf{M}_{\mathrm{act}}^{p}
\label{eq:Mlink_parallel}
\end{equation}
Pulling back the total and link-side inertias to the serial coordinates yields
\begin{equation}
\mathbf{M}_{p\rightarrow s}(\alpha)
=
\mathbf{J}^T\mathbf{M}_{p}(\alpha)\mathbf{J},
\qquad
\mathbf{M}_{\mathrm{link}}^{p\rightarrow s}(\alpha)
=
\mathbf{J}^T\mathbf{M}_{\mathrm{link}}^{p}(\alpha)\mathbf{J}
\label{eq:Mp_pullback}
\end{equation}
The total inertia of the simulator's serial model is decomposed as
\begin{equation}
\mathbf{M}_{s}(\alpha)
=
\mathbf{M}_{\mathrm{link}}^s(\alpha)
+
\mathbf{M}_{\mathrm{act}}^s
\label{eq:Ms_definition}
\end{equation}
where \(\mathbf{M}_{\mathrm{link}}^s\) and \(\mathbf{M}_{\mathrm{act}}^s\) denote the link-side inertia included in the serial model and the diagonal actuator inertia in the serial coordinates, respectively. Accounting for the redistribution of actuator inertia under the coordinate transformation, the joint-space inertia of the serial model after applying S3N-Act becomes
\begin{equation}
\mathbf{M}_{\mathrm{act,norm}}^s(\alpha)
=
\mathbf{M}_{s}(\alpha)
+
\Delta\mathbf{M}_{\mathrm{act}}
=
\mathbf{M}_{\mathrm{link}}^s(\alpha)
+
\mathbf{M}_{\mathrm{act}}^{p\rightarrow s}.
\label{eq:Mactnorm_definition}
\end{equation}
To avoid double-counting the actuator inertia included in the hardware identification results, the residual linkage-inertia increment is defined as follows.

\begin{equation}
\begin{aligned}
\Delta\mathbf{M}_{\mathrm{link}}(\alpha)
&=
\mathbf{M}_{\mathrm{link}}^{p\rightarrow s}(\alpha)
-
\mathbf{M}_{\mathrm{link}}^s(\alpha)\\
&=
\mathbf{M}_{p\rightarrow s}(\alpha)
-
\mathbf{M}_{\mathrm{act,norm}}^s(\alpha).
\end{aligned}
\label{eq:residual_link_inertia}
\end{equation}

The two algebraically equivalent expressions represent, respectively, the difference between the hardware-identified parallel-link and serial-link inertias and that between the hardware-identified total parallel inertia and the actuator-normalized serial reference inertia. The normalized inertia and damping matrices of S3N-Full are

\begin{equation}
\mathbf{M}_{\mathrm{norm}}^s(\alpha)
=
\mathbf{M}_{s}(\alpha)
+
\Delta\mathbf{M}_{\mathrm{act}}
+
\Delta\mathbf{M}_{\mathrm{link}}(\alpha),
\label{eq:Mnorm_update}
\end{equation}
\begin{equation}
\mathbf{D}_{\mathrm{norm}}^s
=
\mathbf{D}_{s}
+
\Delta\mathbf{D}_{\mathrm{act}}.
\label{eq:Dnorm_update}
\end{equation}
The configuration dependence of the residual linkage inertia
also introduces a velocity-dependent term. Under the
inertia-dominant approximation adopted in this study, this term
is neglected, and only the acceleration-dependent contribution is retained. The resulting approximate dynamics are given by

\begin{equation}
\mathbf{M}_{\mathrm{norm}}^s(\alpha)\ddot{\mathbf{q}}^s+
\mathbf{h}_{s}(\mathbf{q}^s,\dot{\mathbf{q}}^s)+
\mathbf{D}_{\mathrm{norm}}^s\dot{\mathbf{q}}^s+
\mathbf{g}_{s}(\mathbf{q}^s)
=
\mathbf{J}^T\boldsymbol{\tau}^p.
\label{eq:normalized_system_dynamics}
\end{equation}
Eq.~\eqref{eq:normalized_system_dynamics} defines S3N, which incorporates the previously derived inertia and damping correction terms into the equations of motion of the serial simulator. Because the correction scope is limited to inertia and damping, \(\mathbf{g}_{s}\) does not include the additional gravitational contribution of the parallel links omitted from the serial-tree surrogate. S3N-Act applies the increments arising from the coordinate transformation of the actuator inertia and damping, whereas S3N-Full additionally includes the residual linkage-inertia increment.

For implementation, the actuator increments are decomposed into implicit and explicit components, and the following matrices are defined.
\begin{equation}
\mathbf{E}_{h}
=
\begin{bmatrix}
1&0\\
0&0
\end{bmatrix},
\qquad
\mathbf{E}_{x}
=
\begin{bmatrix}
0&1\\
1&0
\end{bmatrix}.
\end{equation}
Thus,
\(\Delta\mathbf{M}_{\mathrm{act}}=\bar J_2(\mathbf{E}_{h}+\mathbf{E}_{x})\) and
\(\Delta\mathbf{D}_{\mathrm{act}}=\bar B_2(\mathbf{E}_{h}+\mathbf{E}_{x})\).
The hip-diagonal terms \(\bar J_2\mathbf{E}_{h}\) and \(\bar B_2\mathbf{E}_{h}\) are implemented using the simulator's armature and viscous-damping parameters, respectively, while the actuator off-diagonal and residual-linkage inertia compensation torques are added to the simulator's torque inputs.

The same normalization formulation is applied to all four legs. Letting \(n\in\{1,2,3,4\}\) denote the leg index,
\begin{equation}
\alpha_n=q_{2,n}^{p}-q_{1,n}^{p}=q_{2,n}^{s}.
\end{equation}
The implementation for each leg is given by
\begin{equation}
\begin{aligned}
&
\left(
\mathbf{M}_{s}(\alpha_{n})
+
\bar J_2\mathbf{E}_{h}
\right)
\ddot{\mathbf{q}}_{n}^{s}
+
\left(
\mathbf{D}_{s}
+
\bar B_2\mathbf{E}_{h}
\right)
\dot{\mathbf{q}}_{n}^{s}\\
&\quad
+
\mathbf{h}_{s,n}
+
\mathbf{g}_{s,n}
=
\mathbf{J}^{T}\boldsymbol{\tau}_{n}^{p}
+
\boldsymbol{\tau}_{\mathrm{act,off},n}^{s}
+
\boldsymbol{\tau}_{\mathrm{res},n}^{s}
\end{aligned}
\label{eq:serial_with_tau_norm}
\end{equation}
The actuator-side inertia and damping compensation torque and the residual-linkage inertia compensation torque are computed as
\begin{equation}
\boldsymbol{\tau}_{\mathrm{act,off},n}^{s}
=
-\bar J_2\mathbf{E}_{x}\ddot{\mathbf{q}}_{n}^{s}
-\bar B_2\mathbf{E}_{x}\dot{\mathbf{q}}_{n}^{s},
\label{eq:tau_act_off}
\end{equation}
\begin{equation}
\boldsymbol{\tau}_{\mathrm{res},n}^{s}
=
-\Delta\mathbf{M}_{\mathrm{link}}(\alpha_{n})
\ddot{\mathbf{q}}_{n}^{s}.
\label{eq:tau_res}
\end{equation}
Here, \(\dot{\mathbf{q}}_{n}^{s}\) and \(\ddot{\mathbf{q}}_{n}^{s}\) are the joint-velocity and joint-acceleration feedback available for compensation-torque computation from the preceding simulation step. These values are used directly without additional filtering. The diagonal armature components are handled internally by the simulator, while only the coupling components induced by the actuator coordinate transformation and the residual-linkage inertia components are added to the simulator input as additional generalized torques.

\section{System Identification for S3N}
\label{sec:id}

This section constructs the S3N correction terms derived in the preceding section from hardware FRFs. First, inertia, damping, and torque delay are identified from the actuator-level torque-to-velocity FRF to obtain the actuator parameters for S3N-Act. Next, the total equivalent inertia of the physical parallel leg is identified from the leg-level multiple-input multiple-output (MIMO) FRF measured under closed-loop posture control, and the residual linkage inertia for S3N-Full is obtained by separating the actuator inertia.

\subsection{Identifying Actuator-Side Inertia, Damping, and Delay}

\begin{figure}
    \centering
    \includegraphics[width=0.9\linewidth]{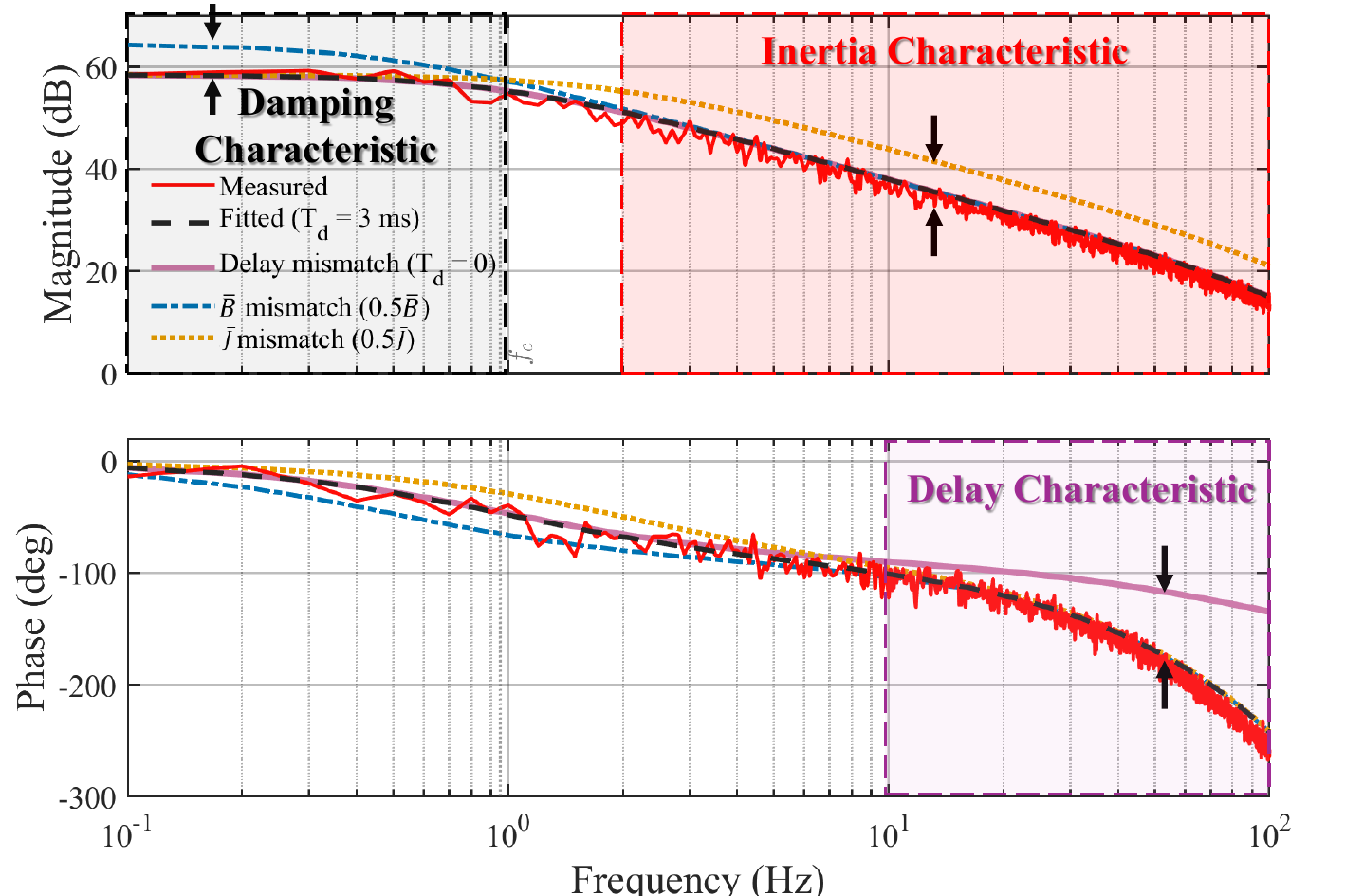}
    \caption{Measured and fitted actuator torque-to-velocity FRFs with parameter-mismatch cases. Damping primarily affects the low-frequency magnitude, inertia the high-frequency roll-off, and command-path delay the frequency-dependent phase lag.}
    \label{fig:Motor_ID_analysis}
\end{figure}

\begin{table}[t]
    \centering
    \caption{Identified actuator-side parameters and common timing
    settings used by all ablation models.}
    \label{tab:actuator_params}
    \footnotesize
    \renewcommand{\arraystretch}{1.08}
    \begin{tabularx}{\columnwidth}
        {>{\raggedright\arraybackslash}Xc}
        \toprule
        Item & Value \\
        \midrule
        Actuator inertia, \(N^2J_m\)
        & \(0.0162~\mathrm{kg\,m^2}\) \\
        Actuator damping, \(N^2B_m\)
        & \(0.0972~\mathrm{N\,m\,s/rad}\) \\
        Command-path delay, \(T_d\)
        & \(3.0~\mathrm{ms}\) \\
        Simulation/control timestep, \(\Delta t_{\mathrm{sim}}\)
        & \(3.0~\mathrm{ms}\) \\
        Actor update period, \(\Delta t_\pi\)
        & \(18.0~\mathrm{ms}\) \\
        \bottomrule
    \end{tabularx}
\end{table}

The actuator inertia, damping, and torque delay were identified
from the actuator torque-to-velocity FRF measured under open-loop
torque excitation using a \(0.1\)--\(100\,\mathrm{Hz}\) Schroeder
multisine
\cite{Schroeder1970Multisine,Pintelon2012FrequencyDomainID}. After removing the transient response and averaging the spectra, the FRF was computed as
\begin{equation}
\hat{G}_a(j\omega_k)
=
\frac{\mathcal{F}\{\dot{q}(t)\}(\omega_k)}
{\mathcal{F}\{\tau(t)\}(\omega_k)}.
\end{equation}

The parameters of the following model were adjusted to match the measured FRF:
\begin{equation}
G_a(j\omega)
=
\frac{e^{-j\omega T_d}}
{\bar{J}j\omega+\bar{B}}.
\label{eq:actuator_frf_model}
\end{equation}

Rather than jointly fitting the time-domain response, we exploit the distinct frequency signatures of each physical component. As shown in Fig.~\ref{fig:Motor_ID_analysis}, damping primarily affects the low-frequency magnitude, inertia the high-frequency roll-off, and torque delay the frequency-dependent phase lag, allowing their effects to be identified separately. The excitation conditions and identification results are summarized in Table~\ref{tab:actuator_params}. The identified \(\bar{J}\) and \(\bar{B}\) are substituted into Eqs.~\eqref{eq:Mact_delta}--\eqref{eq:Dact_delta} to construct the inertia and damping correction terms of S3N-Act, while \(T_d\) is implemented as a one-simulation-step delay in the torque-command path.

\begin{table}[!t]
    \centering
    \caption{Total equivalent inertia in serial coordinates at
    \(\alpha=90^\circ\). Parentheses indicate changes relative to Kin-Only.}
    \label{tab:serial_equiv_inertia_entries}
    \footnotesize

    \renewcommand{\arraystretch}{1.05}
    \begin{tabular}{@{}lccc@{}}
        \toprule
        Method
        & \(M_{11}^s\)
        & \(M_{12}^s\)
        & \(M_{22}^s\) \\
        \midrule
        Kin-Only
        & 0.04058
        & 0.00232
        & 0.01852 \\
        S3N-Act
        & 0.05678 {(+39.9\%)}
        & 0.01852 {(+698.0\%)}
        & 0.01852 {(+0.0\%)} \\
        S3N-Full
        & 0.07161 {(+76.5\%)}
        & 0.02368 {(+920.8\%)}
        & 0.02368 {(+27.9\%)} \\
        \bottomrule
    \end{tabular}
\end{table}


\begin{algorithm}[t]
\caption{Hardware-informed S3N workflow}
\label{alg:s3n_pipeline}
\footnotesize
\SetAlgoLined
\DontPrintSemicolon

\textbf{Step 1: Hardware identification}\;
Acquire actuator-level FRFs $\mathcal{D}_{a}$ and leg-level MIMO FRFs $\mathcal{D}_{\ell}$ from the parallel-link leg\;
Identify actuator parameters $(\bar{J},\,\bar{B},\,T_d)$ from $\mathcal{D}_{a}$ and
total leg inertia $\mathbf{M}_{p}(\alpha)$ from $\mathcal{D}_{\ell}$\;

\BlankLine
\textbf{Step 2: S3N dynamics normalization}\;
\textit{S3N-Act}: Construct actuator-side increments
$\Delta\mathbf{M}_{\mathrm{act}}$,
$\Delta\mathbf{D}_{\mathrm{act}}$
using Eqs.~\eqref{eq:Mact_delta}--\eqref{eq:Dact_delta}\;

\textit{S3N-Full}: Separate the actuator contribution from
$\mathbf{M}_{p}(\alpha)$ and construct
$\Delta\mathbf{M}_{\mathrm{link}}(\alpha)$ using Eq.~\eqref{eq:residual_link_inertia}\;

Inject the increments as compensation torques
($\boldsymbol{\tau}^{s}_{\mathrm{act,off}}$, $\boldsymbol{\tau}^{s}_{\mathrm{res}}$)
into the serial-tree simulator at each timestep\;

\BlankLine
\textbf{Step 3: Policy training}\;
Train policy $\pi_\theta$ via PPO in the S3N-normalized simulator $\mathcal{S}_{\mathrm{S3N}}$\;

\BlankLine
\textbf{Step 4: Hardware deployment}\;
Deploy $\pi_{\theta^\star}$ directly on the parallel-link hardware\;
\textit{No S3N compensation torques are required at deployment}\;

\end{algorithm}

\subsection{Identifying the Full Parallel-Link Lumped Inertia}

The actuator-level FRF is used to identify the actuator inertia, damping, and delay. In contrast, the leg-level MIMO FRF captures the combined inertia of the actuators and linkage, from which the total inertia of the physical parallel-link leg is identified. The previously identified actuator inertia is then separated to construct the residual linkage inertia for S3N-Full.

Unlike actuator identification, the leg-level MIMO identification was performed under joint closed-loop feedback within a safe motion range. To obtain and fit the responses associated with the off-diagonal link-inertia terms, the feedback controller maintained the joint posture at \(\alpha^*=\frac{\pi}{3}\), while the excitation was injected as feedforward torque. To make the input matrix full rank, the input spectrum for each sum- and difference-mode record was defined using the actual total applied actuator torque, computed as the sum of the feedback and injected multisine torques. For excitation mode \(\ell\in\{+,-\}\),
\begin{subequations}
\label{eq:mimo_total_torque_frf}

\begin{equation}
\begin{aligned}
\mathbf{U}^{(\ell)}(j\omega)
&=
\mathcal{F}\!\left\{
\boldsymbol{\tau}_{\mathrm{fb}}^{p,(\ell)}(t)
+
\boldsymbol{\tau}_{\mathrm{inj}}^{p,(\ell)}(t)
\right\},\\
\mathbf{Y}^{(\ell)}(j\omega)
&=
\mathcal{F}\!\left\{
\dot{\mathbf{q}}^{p,(\ell)}(t)
\right\},
\end{aligned}
\label{eq:mimo_mode_spectra}
\end{equation}

\begin{equation}
\begin{aligned}
\mathbf{U}
&=
\begin{bmatrix}
\mathbf{U}^{(+)} & \mathbf{U}^{(-)}
\end{bmatrix},
\qquad
\mathbf{Y}
=
\begin{bmatrix}
\mathbf{Y}^{(+)} & \mathbf{Y}^{(-)}
\end{bmatrix},\\
\widehat{\mathbf{G}}_p
&=
\mathbf{Y}\mathbf{U}^{-1}.
\end{aligned}
\label{eq:mimo_frf_matrix}
\end{equation}

\end{subequations}
Thus, \(\mathbf{U}\) contains the total applied actuator torque rather than the excitation alone, and \(\widehat{\mathbf{G}}_p\) represents the local total torque-to-velocity response around the controlled posture. This is consistent with the frozen-parameter approach used for LPV frequency-domain identification under closed-loop MIMO conditions \cite{Saupe2015MSSP,VanDerMaas2017LPVFRF}. 

\begin{figure}[!t]
    \centering
    \includegraphics[width=0.85\columnwidth]{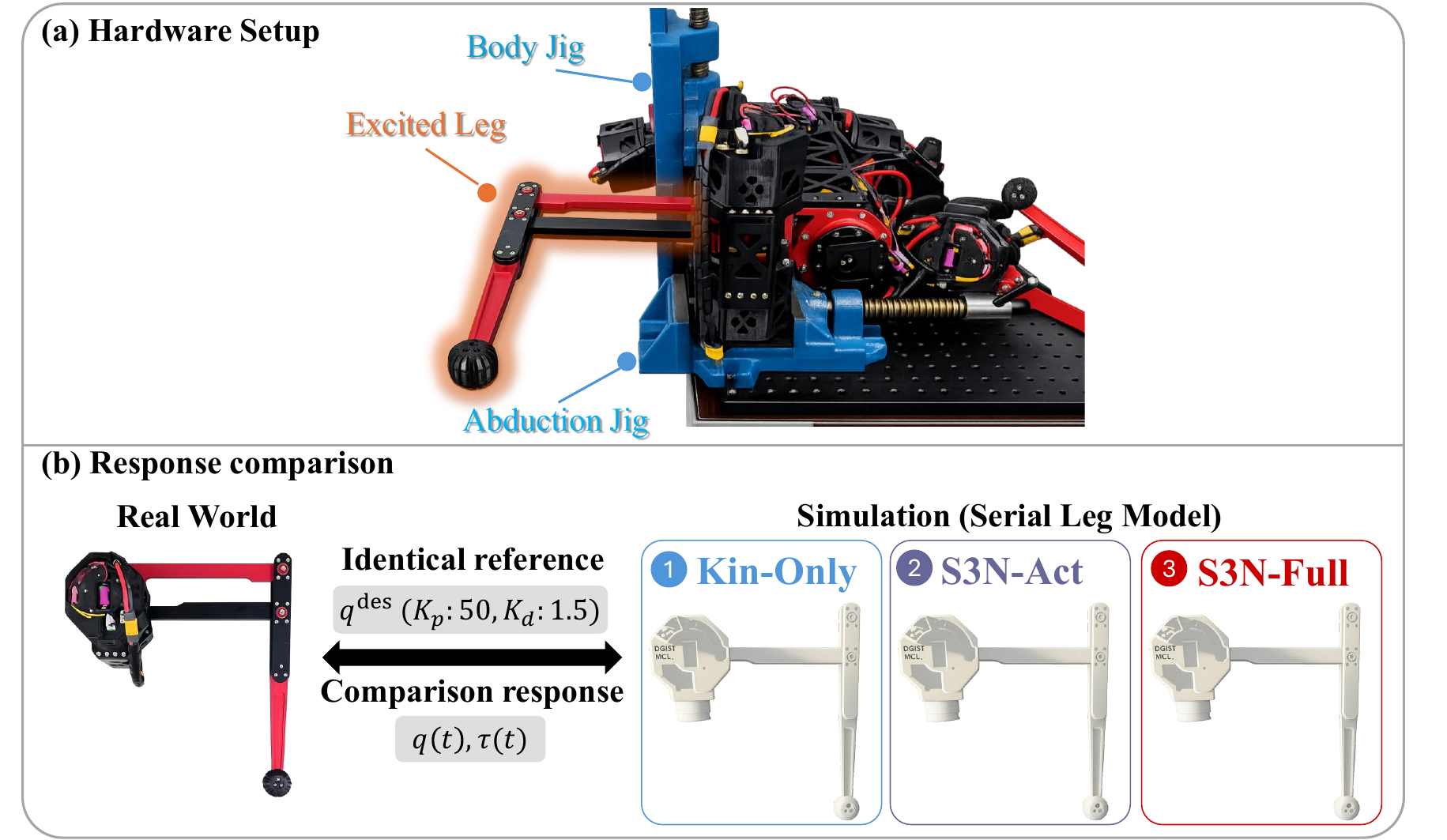}
    \caption{Contact-free dynamic-response validation setup for a 2-DoF parallel-link leg, with all motions other than the two target DOFs mechanically constrained and the two joints excited individually.}
    \label{fig:2dof_setup}
\end{figure}

The identified parameters are
\begin{equation}
\begin{aligned}
M_{11}&=0.04793,\qquad M_{12}=0.005413,\\
M_{22}&=0.02368.
\end{aligned}
\end{equation}
The resulting total inertia of the physical parallel-link leg is
\begin{equation}
\mathbf{M}_p(\alpha)=
\begin{bmatrix}
0.04793&0.005413\cos\alpha\\
0.005413\cos\alpha&0.02368
\end{bmatrix}\;[\mathrm{kg\,m^2}]
\label{eq:identified_inertia_matrix}
\end{equation}
Subtracting the separately identified actuator inertia of \(0.0162\,\mathrm{kg\,m^2}\) from each diagonal entry yields the physical parallel-link inertia matrix
\begin{equation}
\mathbf{M}_{\mathrm{link}}^p(\alpha)=
\begin{bmatrix}
0.03173&0.005413\cos\alpha\\
0.005413\cos\alpha&0.00748
\end{bmatrix}\;[\mathrm{kg\,m^2}]
\label{eq:identified_link_inertia_matrix}
\end{equation}
For S3N-Full, the identified link inertia in Eq.~\eqref{eq:identified_link_inertia_matrix} is transformed into the serial coordinates and substituted into Eq.~\eqref{eq:residual_link_inertia} to construct \(\Delta\mathbf{M}_{\mathrm{link}}(\alpha)\), the difference from the serial-link model. This difference is incorporated through the compensation torque in Eq.~\eqref{eq:tau_res}.

To quantify the dynamics changes induced by normalization, Table~\ref{tab:serial_equiv_inertia_entries} lists the total equivalent inertia entries at \(\alpha=90^\circ\), a representative walking posture. Kin-Only and S3N-Act use the same CAD-based serial-link inertia, while S3N-Act additionally accounts for the actuator-side coordinate-transformation effect. S3N-Full further adds the residual linkage-inertia increment, replacing the serial-link inertia with the hardware-identified parallel-link inertia. Consequently, both the diagonal and coupling terms change substantially under S3N-Act and further under S3N-Full. This represents a substantial mismatch in the inertia terms governing the leg dynamics rather than a minor parameter correction, and can directly alter both robot motion and contact forces by changing the acceleration response to the same torque input.

Algorithm~\ref{alg:s3n_pipeline} summarizes the complete procedure from hardware identification and S3N simulator construction to policy learning and hardware deployment. Identification is performed once before training, and the identified parameters remain fixed during learning. The resulting sim-to-real discrepancies are evaluated in Sections~\ref{sec:mech_validation} and~\ref{sec:quad_eval}.

\section{Contact-Free Validation of the 2-DoF Leg Dynamic Response}
\label{sec:mech_validation}

\begin{figure}[!t]
    \centering
    \includegraphics[width=0.9\linewidth]{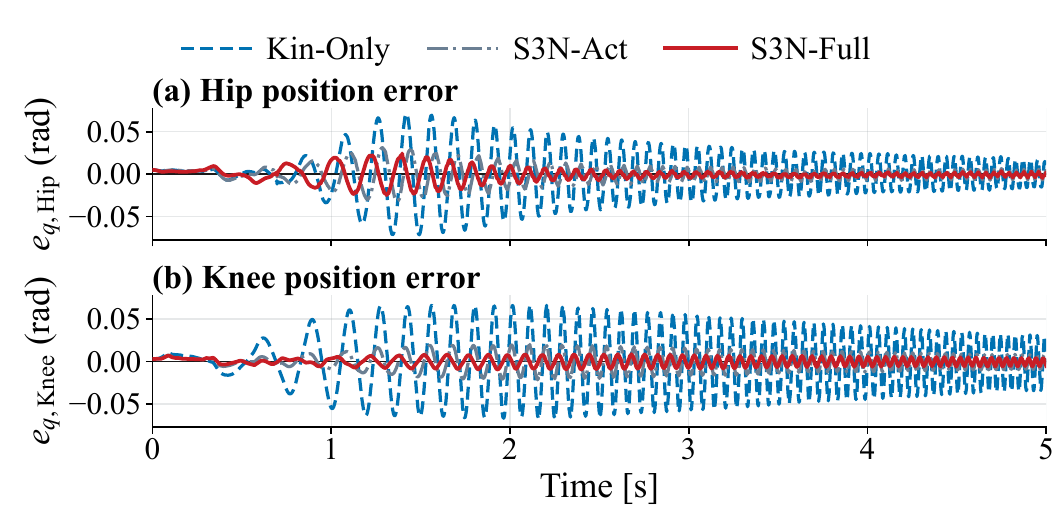}
    \caption{Simulation-to-hardware joint-position error, \(e_q=q_{\mathrm{sim}}-q_{\mathrm{hw}}\).}
    \label{fig:2DOF_position_gap}
\end{figure}

\begin{figure}[!t]
    \centering
    \includegraphics[width=0.9\linewidth]{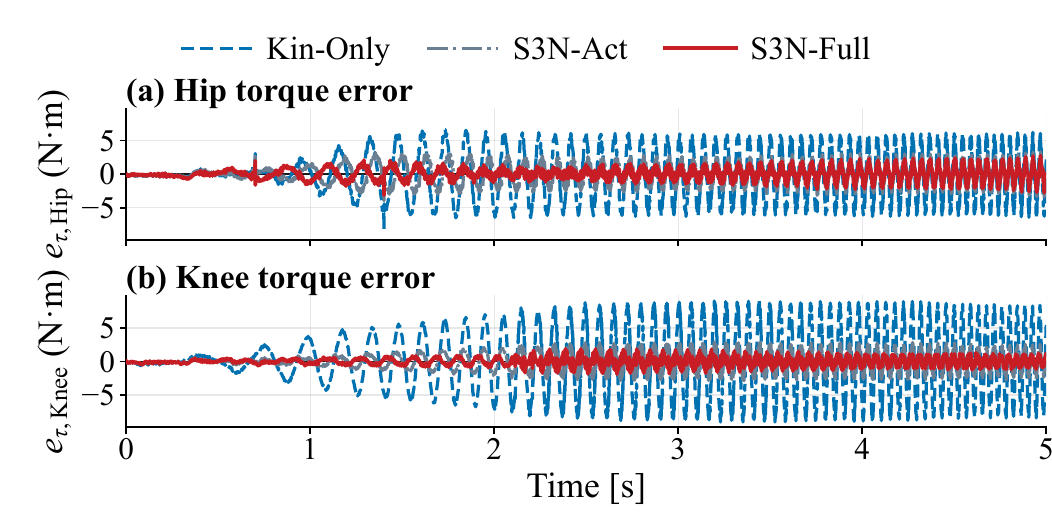}
    \caption{Simulation-to-hardware joint-torque error, \(e_{\tau}=\tau_{\mathrm{sim}}-\tau_{\mathrm{hw}}\).}
    \label{fig:2DOF_torque_gap}
\end{figure}

\begin{figure}[!t]
    \centering
    \includegraphics[width=1\linewidth]{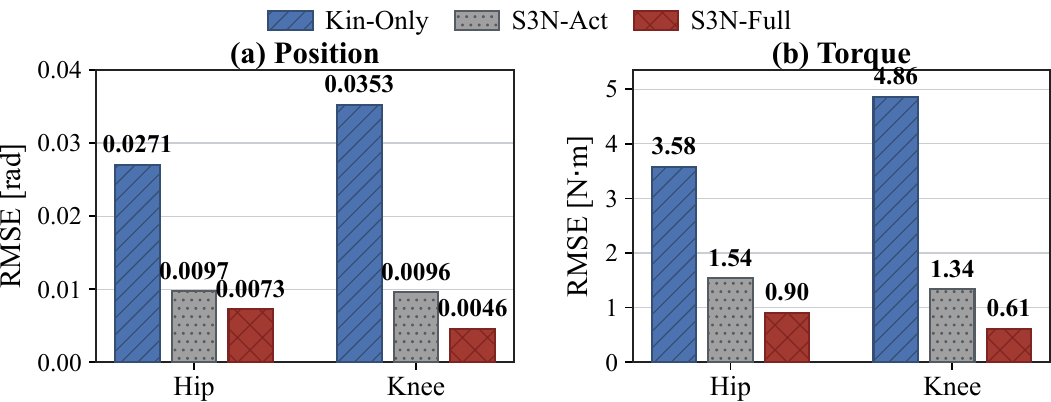}
    \caption{Summary of joint-position and joint-torque RMSEs for the 2-DoF leg dynamic responses. The same reference trajectory was applied to all models.}
    \label{fig:2dof_rmse}
\end{figure}

Before applying reinforcement learning, this section validates the simulation--hardware agreement of the 2-DoF leg dynamics. By comparing the responses of the three simulation models with hardware in the absence of learning and contact, we isolate and evaluate the discrepancies caused by leg-dynamics modeling.

In the validation setup of Fig.~\ref{fig:2dof_setup}, the same reference trajectory was applied to the hardware and all three ablation models. The reference was a desired joint-position chirp with an amplitude of \(0.2\,\mathrm{rad}\) over \(0.1\)--\(25\,\mathrm{Hz}\), and the PD gains were set to \(K_p=50\) and \(K_d=1.5\), as used later for policy learning and hardware evaluation. To isolate the effect of dynamics normalization, all models used the same identified actuator parameters and \(T_d=3.0\,\mathrm{ms}\). Kin-Only uses only the Jacobian-based coordinate and torque mappings, S3N-Act additionally includes \(\Delta\mathbf{M}_{\mathrm{act}}\) and \(\Delta\mathbf{D}_{\mathrm{act}}\), and S3N-Full further includes \(\Delta\mathbf{M}_{\mathrm{link}}\). The joint-position and torque responses were compared.

Figs.~\ref{fig:2DOF_position_gap}--\ref{fig:2dof_rmse} present the simulation--hardware joint-position and torque response errors. Kin-Only exhibited large periodic errors throughout the excitation, whereas the errors of the S3N variants remained relatively close to zero. S3N-Full reduced the hip/knee position RMSEs from \(0.0271/0.0353~\mathrm{rad}\) to \(0.0073/0.0046~\mathrm{rad}\), and the torque RMSEs from \(3.58/4.86~\mathrm{N\!\cdot\!m}\) to \(0.90/0.61~\mathrm{N\!\cdot\!m}\). Averaged over the two joints, the position and torque RMSEs decreased by \(80.9\%\) and \(82.1\%\), respectively, relative to Kin-Only. The joint torques were computed from the measured motor currents and torque constants. Such joint-level dynamics mismatches can lead to sim-to-real discrepancies in quadruped base motion and contact forces, particularly the ground reaction forces.

\section{On-Ground Quadruped Policy Transfer Evaluation}
\label{sec:quad_eval}

This section evaluates whether the improved 2-DoF leg dynamics translate to more accurate GRF responses and sim-to-real transfer of learned quadruped policies. In the pitch-in-place experiment, the rear-right GRF norm is compared between simulation and hardware under a learned policy. In the circular-locomotion experiment, a learned locomotion policy is transferred to hardware, and the linear-velocity and yaw-rate gaps are evaluated.

\begin{figure}
    \centering
    \includegraphics[width=0.8\linewidth]{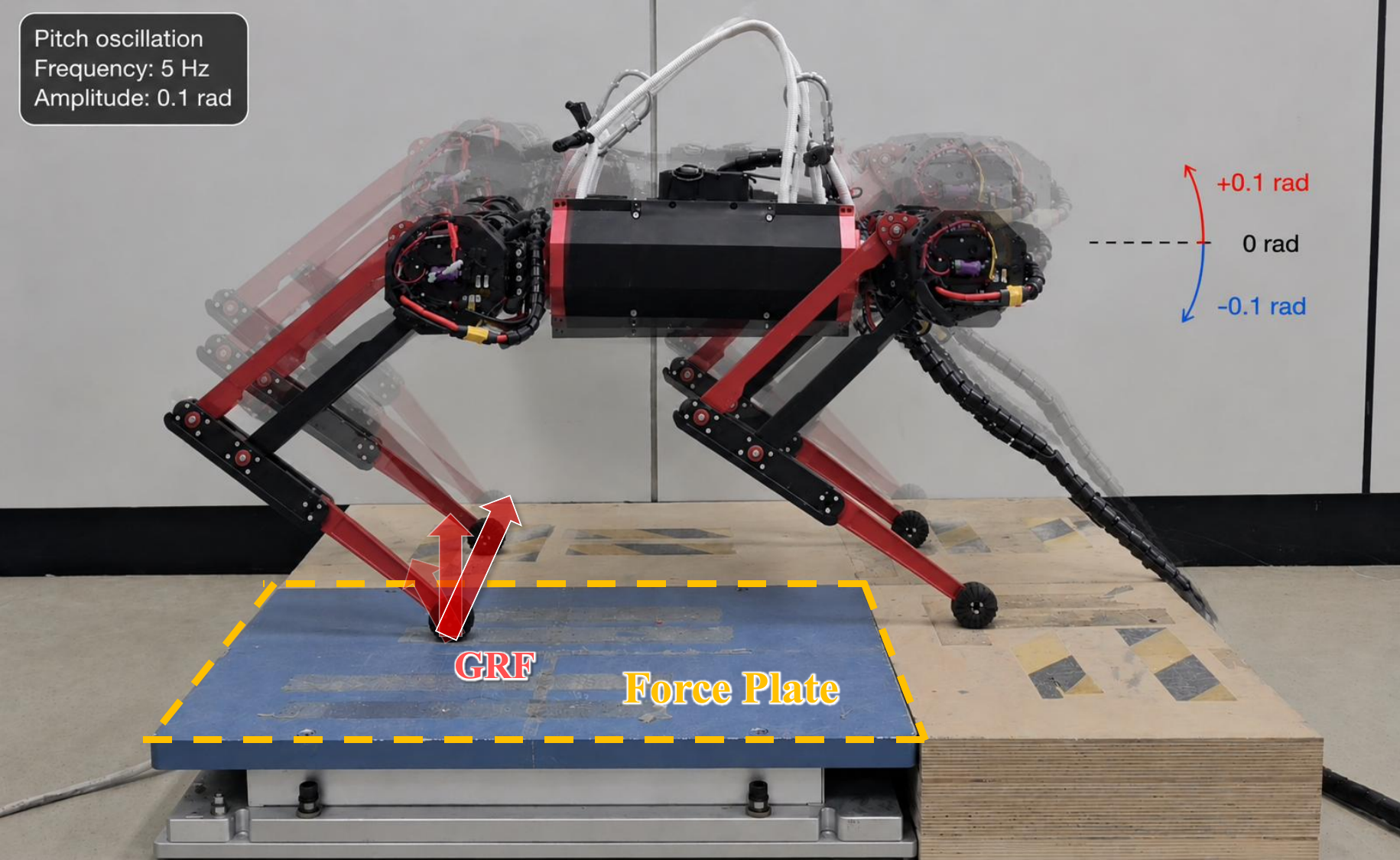}
    \caption{Pitch-in-place force-plate setup. The rear-right foot is placed on the force plate while the learned policy tracks a sinusoidal base-pitch command.}
    \label{fig:pitch_setup}
\end{figure}

\subsection{Evaluation Protocols for Contact-Force Fidelity and Policy Transfer}

For each task, policies were trained independently and evaluated with separate rollouts. Within each task, all settings except the simulator model and leg-link randomization were held fixed, including the rewards, observation and action spaces, network architecture, PPO settings~\cite{Schulman2017PPO}, command distribution, number of parallel environments, low-level control parameters, and number of training iterations. The evaluated methods were Kin-Only, S3N-Act, S3N-Full, and Kin-Only HR. Except for Kin-Only HR, the No Leg-Rand and Leg-Rand
settings in Table~\ref{tab:domain_randomization} were used
for the pitch-in-place and circular-locomotion tasks,
respectively. Kin-Only HR used Heavy Leg-Rand for both
tasks. All methods used identical contact and base
randomization. Kin-Only HR serves as a baseline for testing whether broad nonstructural parameter randomization can reproduce the effect of S3N's structure-aware dynamics normalization.

\begin{figure}
    \centering
    \includegraphics[width=0.75\linewidth]{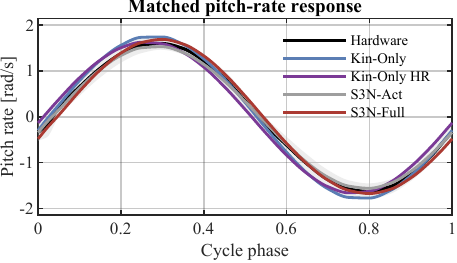}
    \caption{Comparison of closed-loop pitch-rate responses. Each hardware curve represents one cycle obtained by phase-averaging approximately 40 cycles from a single recording. The black curve and shaded region denote the mean and \(\pm1\) standard deviation across the four method-specific hardware responses, while the colored curves show the corresponding nominal simulation rollouts phase-averaged in the same manner.}
    \label{fig:fig_pitch_motion_consistency}
\end{figure}

\begin{figure}
    \centering
    \includegraphics[width=0.75\linewidth]{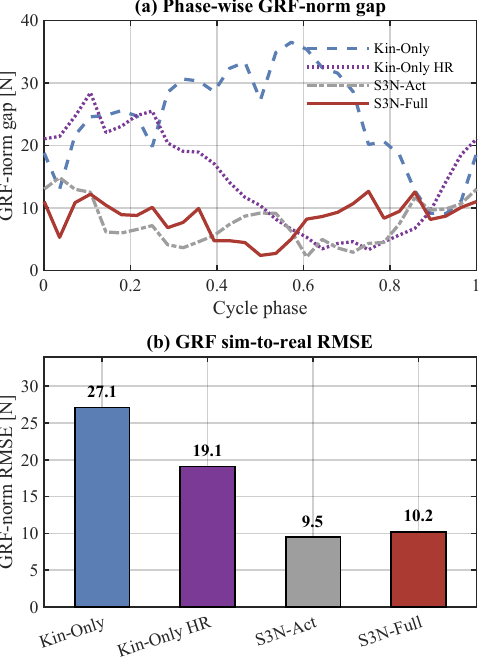}
    \caption{Rear-right GRF-norm sim-to-real gap under similar closed-loop pitch motions. (a) Phase-wise absolute GRF gap, \(\lvert F_{\mathrm{hw}}-F_{\mathrm{sim}}\rvert\), over one cycle. (b) GRF-norm RMSE over the full recording.}
    \label{fig:fig_pitch_grf_validation_vertical_magnitude}
\end{figure}

\begin{table}[!t]
    \centering
    \caption{Leg-link randomization ranges used in the quadruped evaluations.}
    \label{tab:domain_randomization}

    \footnotesize
    \renewcommand{\arraystretch}{1.12}

    \begin{tabular}{@{}lccc@{}}
        \toprule
        Setting
        & \makecell{Leg CoM\\offset [m]}
        & \makecell{Leg inertia\\scale}
        & \makecell{Leg mass\\scale} \\
        \midrule
        No Leg-Rand    & --            & --           & -- \\
        Leg-Rand       & \(\pm0.01\)   & [0.95, 1.05] & [0.95, 1.05] \\
        Heavy Leg-Rand & \(\pm0.03\)   & [0.50, 1.50] & [0.70, 1.30] \\
        \bottomrule
    \end{tabular}
\end{table}

\subsection{Pitch-in-Place: Motion-Matched GRF Validation}
\label{sec:pitch_in_place}

For each method, the pitch-in-place policy was trained with PPO for 1,500 iterations using 4,096 parallel environments. The pitch-in-place experiment was designed to diagnose simulation--hardware dynamics agreement under a fixed-contact stance. As shown in Fig.~\ref{fig:pitch_setup}, the rear-right foot was placed on a force plate, and the same sinusoidal base-pitch reference with a frequency of \(5\,\mathrm{Hz}\) and an amplitude of \(0.1\,\mathrm{rad}\) was applied in simulation and hardware for all methods. The rear-right GRF was measured directly using the force plate on hardware, while the corresponding foot-contact force was extracted in simulation. Except for Kin-Only HR, all methods used the No Leg-Rand setting in Table~\ref{tab:domain_randomization} to exclude the effects of leg-link randomization and isolate differences in how inertia and damping are represented in the simulator dynamics. By eliminating the contact transitions and foot-position variability present during locomotion, we compared the simulation--hardware discrepancies in pitch rate and rear-right GRF norm under similar closed-loop pitch motions.

Fig.~\ref{fig:fig_pitch_motion_consistency} shows similar closed-loop pitch-rate profiles across all methods. The simulation--hardware pitch-rate RMSE, normalized by the measured hardware peak-to-peak range of \(3.232~\mathrm{rad/s}\), was \(3.95\%\), \(4.92\%\), \(1.62\%\), and \(2.64\%\) for Kin-Only, Kin-Only HR, S3N-Act, and S3N-Full, respectively, remaining below \(5\%\) in all cases. In contrast, the rear-right GRF-norm sim-to-real gap in Fig.~\ref{fig:fig_pitch_grf_validation_vertical_magnitude} varied substantially across methods. S3N-Act and S3N-Full showed smaller absolute GRF gaps than Kin-Only over most of the cycle, reducing the GRF-norm RMSE from \(27.1\,\mathrm{N}\) to \(9.5\,\mathrm{N}\) and \(10.2\,\mathrm{N}\), respectively. These correspond to reductions of \(65.1\%\) and \(62.4\%\). These results show that similar closed-loop body motions do not guarantee GRF agreement, as the contact forces required to generate them can vary substantially with the simulator dynamics. 

\begin{figure*}[!t]
    \centering
    \IfFileExists{Figure/fig_circular_sim_real_gap_phase_radius.pdf}{%
        \includegraphics[width=0.85\linewidth]{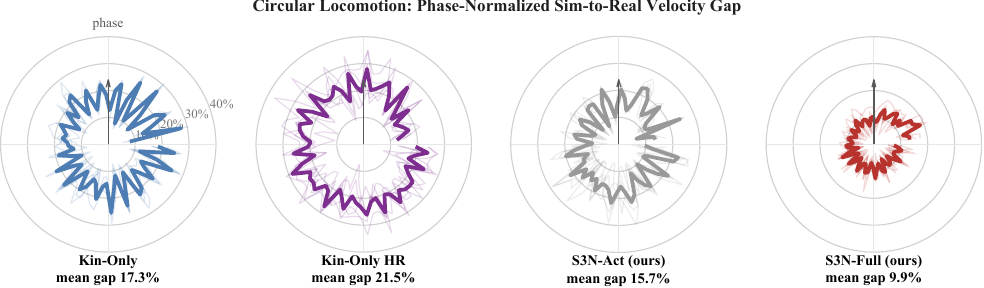}%
    }{%
        \figplaceholder{Figure/fig\_circular\_sim\_real\_gap\_phase\_radius.pdf}%
    }
    \caption{Phase-aligned, command-normalized sim-to-real velocity gaps during circular locomotion. The angular coordinate denotes the circular-command phase, and the radial coordinate denotes the local absolute gap defined in Eq.~\eqref{eq:circular_phase_local_gap}. The thick and light curves represent the trial mean and individual trials, respectively. The value below each panel denotes the cycle-averaged gap \(E_{\phi}\).}
    \label{fig:circular_sim_real_gap_phase}
\end{figure*}

\subsection{Circular Locomotion: Sim-to-Real Velocity Gap}
\label{sec:circular_locomotion}

The circular-locomotion experiment evaluates sim-to-real policy transfer under dynamic conditions involving repeated contact transitions and sustained body translation and rotation. Because the preceding frequency-response analysis showed that inertia mismatch becomes more pronounced at higher frequencies, its effects may be more evident during locomotion with rapid joint accelerations and contact transitions. To account for link-parameter variations caused by manufacturing and assembly tolerances, all methods except Kin-Only HR used the Leg-Rand setting in Table~\ref{tab:domain_randomization}. This differs from the pitch-in-place experiment, which used No Leg-Rand to isolate the effect of simulator dynamics on GRF discrepancies under similar body motions. All circular-locomotion policies were trained for 4,000 PPO iterations using 4,096 parallel environments. To account for trial-to-trial variability, each method was evaluated over five hardware trials, with one lap defined as a trial. We report the mean and standard deviation of each velocity metric.

\begin{table}[!b]
    \centering
    \caption{Sim-to-real velocity metrics for circular locomotion (mean \(\pm\) sample standard deviation over five hardware trials).}
    \label{tab:circular_sim_real_gap}
    \footnotesize
    \renewcommand{\arraystretch}{1.2}
    \begin{tabular}{lcccc}
        \toprule
        Method &
        \makecell{$e_{v_x}$\\$[\mathrm{m/s}]$} &
        \makecell{$e_{\omega_z}$\\$[\mathrm{rad/s}]$} &
        \makecell{$E_{\mathrm{RMSE}}$\\$[\%]$} &
        \makecell{$E_{\phi}$\\$[\%]$} \\
        \midrule
        Kin-Only &
        0.167$\pm$0.006 &
        0.144$\pm$0.011 &
        19.46$\pm$0.87 &
        17.3$\pm$1.0 \\
        Kin-Only HR &
        0.220$\pm$0.016 &
        0.167$\pm$0.028 &
        24.17$\pm$2.50 &
        21.5$\pm$2.4 \\
        S3N-Act &
        0.164$\pm$0.031 &
        0.128$\pm$0.009 &
        18.26$\pm$1.47 &
        15.7$\pm$1.2 \\
        S3N-Full &
        \textbf{0.074$\pm$0.010} &
        \textbf{0.122$\pm$0.007} &
        \textbf{12.23$\pm$0.99} &
        \textbf{9.9$\pm$0.9} \\
        \bottomrule
    \end{tabular}
\end{table}

Hardware trial \(i\) and its corresponding simulation rollout were aligned at the onset of the circular command, and the interval through command termination was analyzed. Over this interval, the sim-to-real RMSEs in forward velocity and yaw rate are denoted by \(e_{v_x}^{(i)}\) and \(e_{\omega_z}^{(i)}\), respectively. The auxiliary metric \(E_{\mathrm{RMSE}}^{(i)}\), obtained by normalizing both components by their command magnitudes and averaging them with equal weights, is defined as
\begin{equation}
E_{\mathrm{RMSE}}^{(i)}
=
\frac{100}{2}
\left(
\frac{e_{v_x}^{(i)}}{\left|v_x^{\mathrm{cmd}}\right|}
+
\frac{e_{\omega_z}^{(i)}}{\left|\omega_z^{\mathrm{cmd}}\right|}
\right)
\label{eq:circular_normalized_rmse}
\end{equation}

The primary metric in this section is the phase-wise command-normalized velocity gap shown in Fig.~\ref{fig:circular_sim_real_gap_phase}. According to the progress phase \(\phi\) of the circular command, one cycle was divided into \(N_{\phi}=72\) equally spaced phase bins centered at \(\phi_j\). Each bin therefore has a width of \(\Delta\phi=2\pi/N_{\phi}=5^\circ\). Within each phase bin, the absolute forward-velocity and yaw-rate gaps were computed from paired simulation--hardware samples and averaged as follows:

\begin{equation}
\begin{aligned}
\epsilon_{\phi}^{(i)}(\phi_j)
=
\frac{100}{2N_{i,j}}
\sum_{k=1}^{N_{i,j}}
\Bigg(
&
\frac{
\left|
v_x^{\mathrm{sim}}(t_k)
-
v_{x,i}^{\mathrm{hw}}(t_k)
\right|
}{
\left|v_x^{\mathrm{cmd}}\right|
}
\\
&+
\frac{
\left|
\omega_z^{\mathrm{sim}}(t_k)
-
\omega_{z,i}^{\mathrm{hw}}(t_k)
\right|
}{
\left|\omega_z^{\mathrm{cmd}}\right|
}
\Bigg).
\end{aligned}
\label{eq:circular_phase_local_gap}
\end{equation}
Here, \(N_{i,j}\) is the number of paired simulation--hardware samples in the \(j\)th phase bin of trial \(i\), and \(t_k\) is the time of the \(k\)th paired sample in that bin. The normalization denominators for \(E_{\mathrm{RMSE}}^{(i)}\) and \(\epsilon_{\phi}^{(i)}\) were set to \(\left|v_x^{\mathrm{cmd}}\right|=0.8\,\mathrm{m/s}\) and \(\left|\omega_z^{\mathrm{cmd}}\right|=0.8\,\mathrm{rad/s}\), respectively. With \(N_{\mathrm{tr}}=5\), the phase-averaged gap \(E_{\phi}^{(i)}\) for trial \(i\) and the overall trial mean \(E_{\phi}\) are computed as follows:

\begin{equation}
E_{\phi}^{(i)}
=
\frac{1}{N_{\phi}}
\sum_{j=1}^{N_{\phi}}
\epsilon_{\phi}^{(i)}(\phi_j),
\qquad
E_{\phi}
=
\frac{1}{N_{\mathrm{tr}}}
\sum_{i=1}^{N_{\mathrm{tr}}}
E_{\phi}^{(i)}
\label{eq:circular_phase_aggregation}
\end{equation}

As shown in Fig.~\ref{fig:circular_sim_real_gap_phase} and Table~\ref{tab:circular_sim_real_gap}, S3N-Full reduced \(E_{\phi}\) and the forward-velocity RMSE by \(42.8\%\) and \(55.7\%\), respectively, relative to Kin-Only, and reduced \(E_{\phi}\) by \(36.9\%\) relative to S3N-Act. It also achieved the lowest RMSE-based auxiliary metric, \(E_{\mathrm{RMSE}}=12.23\pm0.99\%\). Under the tested settings, Kin-Only HR exhibited larger velocity gaps than Kin-Only, indicating that broad unstructured link-parameter randomization alone, within the tested ranges, did not reduce the sim-to-real velocity gap.

\section{Discussion}

\subsection{Structured Dynamics Normalization versus Broad Randomization}

In the pitch-in-place task, Kin-Only HR reduced the GRF sim-to-real gap relative to Kin-Only but was less effective than S3N. During circular locomotion, it exhibited a larger velocity gap than Kin-Only. By contrast, S3N-Full achieved the lowest phase-averaged velocity gap and forward-velocity RMSE. These results show that, under the tested settings, broad parameter randomization alone did not reproduce the advantages of S3N’s structured dynamics normalization.

S3N uses identified parameters to restore the redistribution and coupling of actuator-side inertia and damping, as well as residual linkage inertia, in the simulation dynamics. This restoration reduces systematic model mismatch. Given these distinct roles, a suitable strategy is to first use S3N to construct hardware-consistent nominal dynamics and then apply domain randomization to address residual uncertainty. This strategy is also consistent with prior studies on the limitations of excessive randomization and the adaptation of randomization distributions using data from hardware rollouts \cite{Tan2018SimToReal,Sheckells2019DataDrivenDR,Tiboni2024DORAEMON,Chebotar2019SimOpt}.

\subsection{Scope and Limitations}

S3N is designed for TBCM workflows in which parallel-link mechanisms are represented by serial-tree surrogates. The same Isaac-based serial-tree backend was used for all methods under comparison, and hardware-identified inertia and damping parameters were incorporated into S3N. The effectiveness of S3N was evaluated in terms of policy-level sim-to-real consistency.

However, the Coriolis, centrifugal, and gravitational terms associated with the parallel links omitted during serial-tree reduction were not included. Nonlinear friction \cite{Yeo2025MystericNet}, backlash, compliance, and contact deformation were also excluded from the scope of normalization.

\section{Conclusion}

This paper addressed the sim-to-real gap in dynamics arising when parallel-link legs are modeled as serial-tree surrogates and proposed Simulator-Side System Normalization (S3N) to reduce this gap. Jacobian-based state and torque mappings alone cannot represent the redistribution and coupling of parallel-actuator inertia and damping or the linkage inertia omitted during serial-tree reduction. This study formalized the resulting dynamics gap. In addition, actuator- and leg-level FRFs were identified separately to distinguish the actuator contribution from the residual linkage inertia without double counting, and both components were incorporated into the joint-space dynamics of the serial-tree simulator.

The evaluation covered the position and torque responses of a contact-free 2-DoF parallel leg, the GRF response of the rear-right leg during pitch-in-place motion, and body-velocity consistency during circular locomotion. Relative to Kin-Only, S3N improved sim-to-real consistency across all three evaluations. In particular, the finding that GRF consistency can differ substantially even when body motions are similar shows that motion-level agreement alone does not guarantee force-level fidelity. In addition, S3N preserves the existing serial-tree learning pipeline without explicit loop-closure constraints and exposes the policy to plant dynamics closer to those of the physical parallel-link mechanism. Future work will identify and incorporate additional dynamic components currently excluded from the scope of normalization and extend S3N to a broader range of parallel mechanisms.

\bibliographystyle{IEEEtran}
\bibliography{references}

\end{document}